\documentclass[11pt]{article}
\usepackage{booktabs}
\usepackage{acl}
\usepackage{multirow}
\usepackage{times}
\usepackage{latexsym}
\usepackage{graphicx}
\usepackage{amsmath, amssymb, mathtools}
\usepackage[T1]{fontenc}

\usepackage[utf8]{inputenc}

\usepackage{microtype}

\title{Alleviating Over-smoothing for Unsupervised Sentence Representation }

\author{Nuo Chen\textsuperscript{1},
  Linjun Shou\textsuperscript{2},
  Ming Gong\textsuperscript{2},
  Jian Pei\textsuperscript{3},
  Bowen Cao\textsuperscript{4}\\
  \textbf{
  Jianhui Chang\textsuperscript{4},
  Daxin Jiang\textsuperscript{2},  Jia Li\textsuperscript{1}\thanks{\; Corresponding Author}} \\
  \textit{}
    \textsuperscript{\rm 1}Hong Kong University of Science and Technology (Guangzhou),\\ Hong Kong University of Science and Technology \\
	\textsuperscript{\rm 2}STCA, Microsoft, Beijing,
	\textsuperscript{\rm 3}Duke University, USA\\
 \textsuperscript{\rm 4}Peking University, China\\
\texttt{chennuo26@gmail.com}, \texttt{jialee@ust.hk} \\}

\begin{document}
\maketitle
\begin{abstract}
Currently, learning better unsupervised sentence representations is  the pursuit of many natural language processing communities. Lots of approaches based on pre-trained language models (PLMs) and contrastive learning have  achieved promising results on this task. Experimentally, we observe that the  over-smoothing problem reduces the capacity of  these powerful PLMs, leading to sub-optimal sentence representations. In this paper, we present a \underline{S}imple method named \underline{S}elf-\underline{C}ontrastive \underline{L}earning (SSCL) to alleviate this issue, which samples negatives from PLMs intermediate layers, improving the quality of the sentence representation. Our proposed method is quite simple and can be easily extended to various state-of-the-art models for performance boosting, which can be seen as a plug-and-play contrastive framework for learning unsupervised sentence representation. Extensive results prove that SSCL brings the superior performance improvements of different strong baselines (e.g., BERT and SimCSE) on Semantic Textual Similarity and Transfer datasets. Our codes are available at \url{https://github.com/nuochenpku/SSCL}.
\end{abstract}

\section{Introduction}
\label{introduction}
Learning effective sentence representations is a long-standing and fundamental goal of natural language processing (NLP) communities \cite{hill-etal-2016-learning,conneau-etal-2017-supervised-infersent,kim2021self,DBLP:journals/corr/abs-2104-08821}, which can be applied to various downstream NLP tasks such as Semantic Textual Similarity \cite{agirre-etal-2012-semeval,agirre-etal-2013-sem,agirre-etal-2014-semeval,agirre-etal-2015-semeval,agirre-etal-2016-semeval,cer-etal-2017-semeval,marelli-etal-2014-sick} and information retrieval \cite{xiong2021approximate}.
Compared with supervised sentence representations, unsupervised sentence representation learning is more challenging because of lacking enough supervised signals.

\begin{figure}[!t]
\centering
\includegraphics[width=0.9\linewidth]{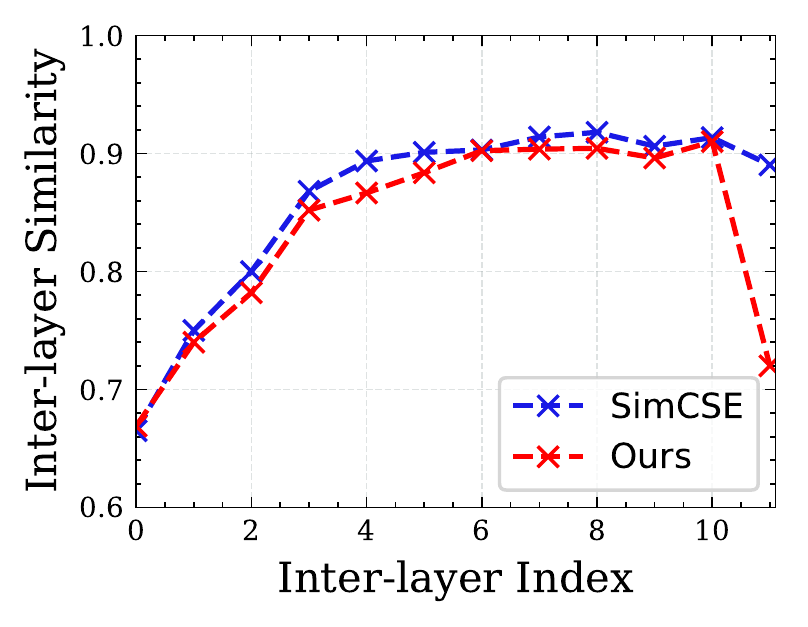}
\vspace{-10pt}
\caption{Inter-layer cosine similarity of sentence representations computed from SimCSE and Ours. We calculate the sentence representation similarity between two adjacent layers on STS-B test set. In this example, we extend our methods of SimCSE by utilizing the penultimate layer as negatives.}
\label{fig:layer_similarity}
\vspace{-10pt}
\end{figure}


In the context of  unsupervised sentence representation learning, prior works \cite{DBLP:journals/corr/abs-1810-04805,DBLP:conf/iclr/LanCGGSS20} tend to directly utilize large pre-trained language models (PLMs) as the sentence encoder to achieve promising results. Recently, researchers point that the representations from these PLMs suffer from  the  anisotropy \cite{li-etal-2020-sentence,su2021whitening} issue, which denotes the learned representations are always distributed into a narrow one in  the semantic space.
More recently, several works \cite{DBLP:conf/acl/GiorgiNWB20, DBLP:journals/corr/abs-2104-08821} prove that incorporating  PLMs with contrastive learning can alleviate this problem, leading to the distribution of sentence representations becoming more uniform. In practice, these works \cite{wu2020clear,yan2021consert} propose various data augmentation methods to construct positive sentence pairs. For instance, \citet{DBLP:journals/corr/abs-2104-08821} propose to leverage dropout as the simple yet effective augmentation method to construct positive pairs, and the corresponding results are better than other more complex augmentation methods. 



Experimentally, aside from the anisotropy and tedious sentence augmentation issues,
we observe a new phenomenon also makes the model sub-optimized: \textit{Sentence representation  between two adjacent layers in the unsupervised sentence encoders are becoming relatively identical when the encoding layers go deeper}. Figure \ref{fig:layer_similarity} shows the sentence representation similarity between two adjacent layers on STS-B test set. The similarity scores in blue dotted line are computed from  SimCSE \cite{DBLP:journals/corr/abs-2104-08821}, which is the state-of-the-art PLM-based sentence model. Obviously, we can observe the similarity between two adjacent layers (inter-layer similarity) is very high (almost more than 0.9). Such high similarities refer to that the model doesn't acquire adequate distinct knowledge as the encoding layer increases, leading to the neural network validity and energy \cite{DBLP:journals/corr/abs-2006-13318} decreased and the loss of discriminative power. In this paper, we call this phenomenon as the \textit{inter-layer} over-smoothing  issue.

\begin{table}[]
\centering
\small
\begin{tabular}{cccc}

\toprule
\textbf{Model}  & SimCSE (10) & SimCSE (12) & \textbf{Ours}\\ \midrule
\textbf{Performance} & 70.45 & 76.85  & \textbf{79.03} \\ \bottomrule
\end{tabular}
\caption{Spearman's correlation score of different models on STS-B. SimCSE (10) and SimCSE (12) means we use the 10 and 12 transformer layers in the encoder.}
\vspace{-10pt}
\label{table11}

\end{table}

Intuitively, there are two factors could result in the above issue: (1) The encoding layers in the model are of some redundancy; (2) The training strategy of current model is sub-optimized, making the deep layers in the encoder cannot be optimized effectively. For the former, the easiest and most reasonable  way is to cut off some layers in the encoder. However, this method inevitably leads to  performance drop. As presented in Table \ref{table11}, the performance of SimCSE decreases from 76.85$\%$ to 70.45$\%$ when we drop the last two encoder layers. 
Meanwhile, almost none existing works have delved deeper to alleviate the over-smoothing issue from the latter side.

Motivated by the above concerns, we present a
 new training paradigm based on contrastive learning:  \underline{S}imple contrastive method named \underline{S}elf-\underline{C}ontrastive \underline{L}earning (SSCL), which can significantly improve the performance of learned sentence representations while alleviating the over-smoothing issue. Simply Said, we utilize hidden representations from intermediate PLMs layers as negative samples  which the final sentence representations should be away from. Generally, our SSCL has several advantages: (1) It is fairly straightforward and does not require complex data augmentation techniques; (2) It can be seen as a contrastive framework that focuses on mining negatives effectively, and can be easily extended into different sentence encoders that aim for building positive pairs; (3) It can further be viewed as a plug-and-play framework for enhancing sentence representations.
As presented in Figure \ref{fig:layer_similarity},  ours (red dotted line) that extend of SimCSE with employing the penultimate layer sentence representation as negatives results
in a large drop in the inter-layer similarity between last two adjacent layers (11-th and 12-th), showing SSCL makes inter-layer sentence representations  more discriminative. Results in Table \ref{table11} show ours also could result in better sentence representations while alleviating the \textit{inter-layer} over-smoothing issue.

We show SSCL brings superior performance improvements in  7 Semantic Textual Similarity (STS) and 7 Transfer (TS) datasets. Experimentally, we apply our method on two base encoders: BERT and SimCSE. And the resulting models achieve 
15.68$\%$ and 1.65$\%$  improvements on STS tasks, separately.
Then, extensive in-depth analysis and probing tasks are further conducted, revealing SSCL could improve PLMs' capability to capture the surface, syntactic and semantic information of sentences via addressing the over-smoothing problem. Besides of these observations, another interesting finding is that ours can keep comparable performance while reducing the sentence vector dimension size significantly\footnote{In real industry scenarios like search, embedding vector dimension is an important factor to influence the dense retrieval serving cost. Larger size means higher serving cost.}. For instance, SSCL even obtains better performances (62.42$\%$ vs. 58.83$\%$) while reducing the  vector dimensions from 768 to 256 dimensions when extending to BERT-base.
In general, the contributions of this paper can be summarized as:
\begin{itemize}
\item We first observe the \textit{inter-layer} over-smoothing issue in current state-of-the-art unsupervised sentence models, and then propose SSCL to alleviate this problem, producing superior sentence representations.
\item Extensive results prove the effectiveness of the proposed SSCL on Semantic Textual Similarity and Transfer datasets.
\item Qualitative and quantitative analysis are included to justify the designed architecture and look into the representation space of SSCL.

\end{itemize}




\section{Background}
In this section, we first review the formulation of the over-smoothing issue in PLMs from the perspective of the \textit{intra-layer} and \textit{inter-layer}. Then we discuss the difference of over-smoothing and annisotropy problems.

\label{background}
\begin{figure*}[!t]
\vspace{-5pt}
\centering
\includegraphics[width=0.9\linewidth]{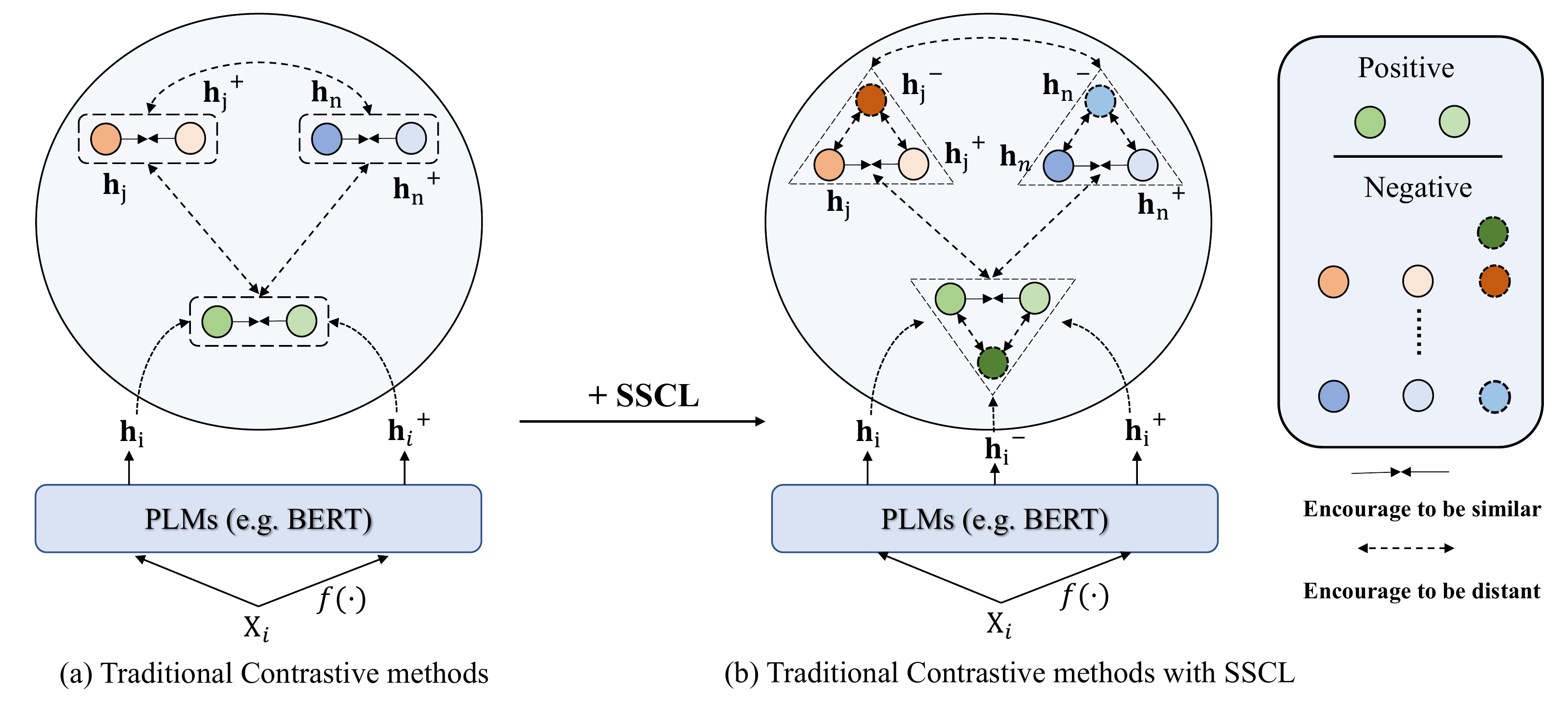}
\caption{The overall architecture of  the traditional contrastive methods and our proposed SSCL.}
\label{fig:framework}

\end{figure*}

\subsection{Over-smoothing}

Recently, \citet{DBLP:journals/corr/abs-2202-08625} point
\textit{intra-layer} over-smoothing issue in PLMs from the perspective of graph, which denotes different tokens in the  input sentence are mapped to quite similar representations. It can be observed 
via measuring the similarity between different tokens in the same sentence, named token-wise cosine similarity. Given a sentence $\texttt{X} = \{\texttt{x}_1,\texttt{x}_2,...,\texttt{x}_m\}$, token-wise cosine similarity of $\texttt{X}$ can be calculated as:
\begin{equation}
    \mathrm{TokSim} =\frac{1}{m(m-1)}\sum_{u\neq v}\frac{\mathbf{x}_u^\top
	 \mathbf{x}_v}{\Vert \mathbf{x}_u\Vert_2\Vert
	 \mathbf{x}_v\Vert_2}
	 \label{token_sim}
\end{equation}
where m is the number of tokens, $\mathbf{x}_u,
\mathbf{x}_v$ are the representations of $\texttt{x}_u,\texttt{x}_v$ from PLMs  and $\Vert\cdot\Vert_2$ is the Euclidean norm.

In this paper, we argue that  the over-smoothing issue also  also exists in \textit{inter-layer} level, which refers to  sentence  representations from adjacent PLMs  layers  are relatively identical. In detail, \textit{inter-layer} over-smoothing means the sentence representations from adjacent layers have high similarity, which can be measured by inter-layer similarity: 
\begin{equation}
    \mathrm{SetSim} =\frac{\mathbf{s}_i^\top
	 \mathbf{s}_{i+1}}{\Vert \mathbf{s}_i\Vert_2\Vert
	 \mathbf{s}_{i+1}\Vert_2}
\end{equation}
where $\mathbf{s}_i$ and $\mathbf{s}_{i+1}$ denote sentence representations of $\texttt{X}$ from two adjacent layers (i-th and i+1-th) in PLMs.

In summary,  the over-smoothing issue can divided into two folds: \textit{inter-layer} and \textit{intra-layer}. In this paper, we aim at alleviating the  over-smoothing issue from the perspective of \textit{inter-layer},
 improving the sentence representations.
Surprisingly, we find alleviating over-smoothing in \textit{inter-layer} also can alleviate the \textit{intra-layer} over-smoothing issue to some extent, which is discussed in Section \ref{discussion_sscl}.

\subsection{Over-smoothing vs. Anisotropy}
Currently, the anisotropy issue  is widely studied  to improve sentence representations from PLMs. Admittedly, despite over-smoothing and anisotropy are related concepts, they are nonetheless completely diverse.
As described in \cite{li-etal-2020-sentence,su2021whitening}, the anisotropy problem refers to the distribution of learnt sentence representations in the semantic space is always constrained to a certain area.
As illustrated in \cite{DBLP:journals/corr/abs-2202-08625}, "over-smoothing” can be summarized as the token uniformity problems in BERT, which denotes token representations in the same input sentence are highly related that is defined as \textit{intra-layer} over-smoothing in this paper. Moreover,  we extend the concept of over-smoothing issue to the \textit{inter-layer}, which refers there is a significant degree of similarity between sentence representations from neighbouring neural network layers.
Experimentally, the over-smoothing problem can cause one sentence to have a greater token-wise similarity or nearby layers  in PLMs to have a higher sentence representation similarity, while anisotropy makes all pairs of sentences in the dataset achieve a relatively identical similarity score.  Obviously, over-smoothing is different from the anisotropy issue. Therefore, we distinguish these two concepts in the paper.

\section{Methodology}
\label{method}
In this section, we first introduce the traditional contrastive methods for learning unsupervised sentence representation. Then, we describe the proposed method SSCL for building negatives and briefly illustrate how to extend SSCL of other contrastive frameworks.

\subsection{Traditional Contrastive Methods}

Considering learning unsupervised sentence representation via contrastive learning needs to construct plausible positives or negatives, traditional contrastive methods (e.g., word deletion, dropout) tends to utilize data augmentation on training data to build positives. In detail, given a sentence collection:  $\mathcal{X}=\{\texttt{X}_i\}_{i=1}^m$.  Subsequently, we can utilize some data augmentation methods: $f(\cdot)$ on each $\texttt{X}_i \in \mathcal{X}$ to construct the semantically related positive sample $\texttt{X}_i^{+} = f(\texttt{X}_i)$ (e.g., dropout, word shuffle and deletion), as shown in Figure \ref{fig:framework} (a). Then, let $\mathbf{h}_i$ and $\mathbf{h}_i^+$ denote the PLMs (e.g., BERT) last layer sentence representations of $\texttt{X}_i$ and $\texttt{X}_i^{+}$, the contrastive training objective for $(\mathbf{h}_i,\mathbf{h}_i^+)$ with a mini-batch of $N$ pairs can be formulated as:
\begin{equation}
   \mathcal{L}_{\texttt{tcm}}= -\texttt{log} \frac{\texttt{exp}(\Psi (\mathbf{h}_i,\mathbf{h}_i^+)/\tau)}{\sum_{j=1}^{N} \texttt{exp} (\Psi (\mathbf{h}_i,\mathbf{h}_j)/\tau)}
\label{eq:contrast}
\end{equation}
where  $\Psi$(,) denotes the cosine similarity function, $\tau$ is temperature. Notice that, these methods focus on mining positive examples while directly utilize in-batch negatives during training. Thereafter, we introduce SSCL to build useful negatives, and thus, can be seen as complementary to previous methods.
\subsection{SSCL}
SSCL is free from external data augmentation procedures which utilizes hidden representations from PLMs intermediate layers as negatives. In this paper, we treat the last layer representation as the final sentence representation which is needed to optimize.
Concretely, we collect the intermediate M-th layer sentence representation in PLMs, which is regarded as the negatives of last layer representation and named as $\mathbf{h}_i^-$, as shown in Figure \ref{fig:framework} (b).
 Hence, we obtain the negative pairs $(\mathbf{h}_{i},\mathbf{h}_i^-)$.  As aforementioned, we also treat $\mathbf{h}_i^+$ as the positive sample which obtained from any data augmentation method. Subsequently, the training objective $\mathcal{L}_{\texttt{hne}}$ can be reformulated as follows:
\begin{equation}
\small
\begin{aligned}
  -\texttt{log} \frac{\texttt{exp}(\Psi (\mathbf{h}_i,\mathbf{h}_i^+)/\tau)}{\sum^{N}( (\texttt{exp} (\Psi (\mathbf{h}_i,\mathbf{h}_j)/\tau) + \texttt{exp} (\Psi (\mathbf{h}_i,\mathbf{h}_i^-)/\tau)) }
\label{eq:simhne}
\end{aligned}
\end{equation}
where the first term in the denominator refers to the original in-batch negatives, and the second term denotes the intermediate negatives. Through these methods, SSCL makes the last layer representation of PLMs more discriminative from the previous layers via easily enlarging the number of negatives, and thus, alleviating the over-smoothing issue. Clearly, our approach is rather straightforward and can be simply implemented into these conventional contrastive techniques.


\section{Experiments}
\label{experiments}
\begin{table*}[t]
    \begin{center}
    \centering
    \small
    \begin{tabular}{lcccccccc}
    \toprule
       $\mathbf{Model}$ &  $\mathbf{STS12}$ & $\mathbf{STS13}$ & $\mathbf{STS14}$ & $\mathbf{STS15}$ & $\mathbf{STS16}$ & $\textbf{STS-B}$ & $\textbf{SICK-R}$ & $\mathbf{Avg.}$ \\
    \midrule
    
        \multicolumn{9}{c}{\it{Base Version}}\\
    \midrule
        GloVe embeddings (avg.)$^\heartsuit$ & 55.14 & 70.66 & 59.73 & 68.25 & 63.66 & 58.02 & 53.76 & 61.32 \\
        \midrule
        BERT~(cls.) & 29.70&	49.38&	39.67&	56.03&	56.19&	43.87&	52.06&	46.70\\ 
        $\textbf{SSCL-BERT}$~(cls.) & 49.21&	\textbf{67.59}&	\textbf{58.96}&	\textbf{69.94}&	68.00&	62.87&	60.43&	62.42\\
        \midrule
        BERT~(avg.) & 48.26&	47.72&	46.83&	52.30&	59.88&	54.27&	56.41&	52.24\\ 
        
        $\textbf{SSCL-BERT}$~(avg.) & \textbf{53.93}&	63.10&	56.41&	68.00&	\textbf{70.46}&	\textbf{64.85}&	\textbf{61.15}&	\textbf{62.56}\\
        \midrule
        BERT-flow$^\heartsuit$ & 58.40&	67.10&	60.85&	75.16&	71.22&	68.66&	64.47&	66.55 \\ 
        BERT-whitening$^\heartsuit$ & 57.83& 66.90 & 60.90 & 75.08& 71.31& 68.24& 63.73& 66.28\\ 
        IS-BERT$^\heartsuit$ & 56.77 & 69.24 & 61.21 & 75.23 & 70.16 & 69.21 & 64.25 & 66.58 \\
        CT-BERT$^\heartsuit$ & 61.63 & 76.80 & 68.47 & 77.50 & 76.48 & 74.31 & 69.19 &72.05 \\
         SimCSE  & 68.40&	82.41 &	74.38&	80.91&	$\mathbf{78.56}$&	76.85&	$\mathbf{72.23}$&	76.25\\
         $\textbf{SSCL-SimCSE}$ & $\mathbf{71.68}$&	$\mathbf{83.50}$ &	$\mathbf{76.42}$&	$\mathbf{83.46}$&78.39&	$\mathbf{79.03}$&71.76&	$\mathbf{77.90}$ \\
        \midrule
         \multicolumn{9}{c}{\it{Large Version}}\\
        \midrule
        BERT$_{large}$ (cls.) & 33.06&	57.64&	47.95&	55.83&	62.42&	49.66&	53.87&	51.49\\
        BERT$_{large}$-flow & 65.20& 73.39&	69.42&	74.92&	77.63&	72.26&	62.50& 70.76\\
        BERT$_{large}$-whitening & 64.35& 74.64&	69.64&	74.68&	75.94&	60.81&	72.47& 70.35\\ 
        Consert$_{large}$   & 70.69& 82.96&	74.13&	82.78&	76.66&	77.53&	70.47& 76.45\\
         SimCSE$_{large}$ & 69.17 & 84.36& 75.09 & 83.99 & 78.61 & 79.54 & 71.97 & 77.53\\
        $\textbf{SSCL-SimCSE}$$_{large}$ & $\mathbf{71.98}$ &$\mathbf{85.74}$	&	$\mathbf{77.94}$	&	$\mathbf{85.94}$	&	$\mathbf{80.08}$	&	$\mathbf{81.20}$	&	$\mathbf{74.28}$ & $\mathbf{79.69}$\\
    \bottomrule
    \end{tabular}
    \end{center}

    \caption{
        Sentence embedding performance on STS tasks (Spearman's correlation, ``all'' setting).
        We highlight the highest numbers among models with the same pre-trained encoder. We run each experiment three times and report average results. $^\heartsuit$ denotes results from \cite{DBLP:journals/corr/abs-2104-08821}.
    }
    \label{tab:main_sts}
    \vspace{-5pt}
\end{table*}
\subsection{Evaluation Datasets}
We conduct our experiments on 7 Semantic Textual Similarity (STS) tasks and 7 Transfer tasks (TR). Following
the  common setting, SentEval toolkit is used for evaluation purposes.

\paragraph{Semantic Textual Similarity} We evaluate our method on the following seven STS datasets: STS 12-16 \cite{agirre-etal-2012-semeval,agirre-etal-2013-sem,agirre-etal-2014-semeval,agirre-etal-2015-semeval,agirre-etal-2016-semeval}, STS-B \cite{cer-etal-2017-semeval} and SICK-R \cite{marelli-etal-2014-sick}. And Spearman’s correlation coefficient is used as evaluation metric of the model performance. 

\paragraph{Transfer} 
We  evaluate our models on the following transfer tasks: MR~\cite{pang2005seeing_mr}, CR~\cite{hu2004mining_cr}, SUBJ~\cite{pang2004sentimental_subj}, MPQA~\cite{wiebe2005annotating_mpqa}, SST-2~\cite{socher2013recursive_sst-2}, TREC~\cite{voorhees2000building_trec} and MRPC~\cite{dolan-brockett-2005-automatically-mrpc}. Concretely, we also follow the default settings in \cite{DBLP:journals/corr/abs-2104-08821} to train each sentence representation learning method. 

\subsection{Implementation Details}
We use the same training corpus from \cite{DBLP:journals/corr/abs-2104-08821} to avoid training bias, which consists of one million sentences randomly sampled from Wikipedia. In our SSCL implement, we select BERT (base and large version) as our backbone architecture because of its typical impact. $\tau$ is set to 0.05 and Adam optimizer is used for optimizing the model. Experimentally, the learning rate is set to 3e-5 and 1e-5 for training BERT$_{base}$ and BERT$_{large}$ models. The batch size is set to 64 and max sequence length is 32. It is worthwhile to notice we utilize average pooling over input sequence token representation and $\texttt{[CLS]}$ vector to obtain sentence-level representations, separately.
More concretely, we train our model with 1 epoch on a single 32G NVIDIA V100 GPU.  For STS tasks, we save our checkpoint with best results on STS-B development set; For Transfer tasks, we use the average score of 7 seven transfer datasets to find the best checkpoint.


\subsection{Results}

\begin{table*}[ht]
    \begin{center}
    \centering
    \small

    \begin{tabular}{lcccccccc}
    \toprule
       \textbf{Model} & \textbf{MR} & \textbf{CR} & \textbf{SUBJ} & \textbf{MPQA} & \textbf{SST} & \textbf{TREC} & \textbf{MRPC} & \textbf{Avg.}\\
    \midrule
        GloVe embeddings (avg.) & 77.25&    78.30&  91.17&  87.85&  80.18&  83.00& 72.87 & 81.52\\
        Skip-thought &  76.50& 80.10&  93.60&  87.10&  82.00&  92.20&  73.00& 83.50  \\
        \midrule
        BERT$_{base}$ (cls.) & 76.86 & 82.68 & 93.73 & 85.87 & 80.56 & 88.20 & 70.13 & 82.57 \\
        \textbf{SSCL-BERT}$_{base}$ (cls.)& 80.48 & \textbf{85.88} & 95.26 & 86.97 & \textbf{84.84} & \textbf{88.80} & 69.62 & \textbf{84.55} \\
        \midrule
        BERT$_{base}$ (avg.) & 77.67& 83.12 & 94.46 & 86.11 & 80.08 & 85.12 & 72.64 & 82.86 \\
        
        \textbf{SSCL-BERT}$_{base}$ (avg.)& \textbf{78.87}& 84.28 & \textbf{95.31} & \textbf{87.40} & 80.79 & 86.00 & 73.12 & 83.68 \\
        \midrule
        SimCSE$_{base}$ & 81.62&	85.44&	94.01&	88.05&	85.06&	89.10&	74.03&	85.11\\
        \textbf{SSCL-SimCSE}$_{base}$ & \textbf{81.08}&	\textbf{86.16}&	\textbf{94.21}&	\textbf{88.63}&	\textbf{85.24}&	\textbf{89.61}&	\textbf{74.20}&	\textbf{85.61}\\
 
     \midrule
        BERT$_{large}$ (cls.) & 78.68 & 84.85 & 94.21 & 88.23 & 84.13 & 91.40 & 71.13 & 84.66 \\
        SSCL-BERT$_{large}$ (cls.)& 73.93 & 87.18 & 94.96 & 88.75 & 85.96 & 88.64 & 74.24 & 85.83 \\
     SimCSE$_{large}$ & 84.37&	88.64&	95.26&	88.04&	89.95&	90.40&	74.42&	87.17\\
    \textbf{SSCL-SimCSE}$_{large}$ & \textbf{86.01}&	\textbf{90.36}&	\textbf{95.98}&	\textbf{89.04}&	
         \textbf{91.27}&	\textbf{93.20}&	\textbf{76.29}& \textbf{88.88}\\
     
    \bottomrule
    \end{tabular}
    \end{center}

    \caption{
        Transfer task results of different sentence embedding models (measured as accuracy). 
        We highlight the highest numbers among models with the same pre-trained encoder.
    }
    \label{tab:main_transfer}
\end{table*}

\paragraph{Baselines} We compare our methods with the following baselines: (1) naive baselines: GloVe average embeddings  \cite{pennington-etal-2014-glove}, Skip-thought and BERT; (2) strong baselines based on BERT: BERT-flow \cite{li-etal-2020-sentence}, BERT-whitening \cite{su2021whitening}, IS-BERT \cite{zhang-etal-2020-unsupervised}, CT-BERT \cite{carlsson2021semantic}, Consert \cite{ DBLP:conf/acl/YanLWZWX20} and SimCSE. For a fair comparison, we extend SSCL to BERT and SimCSE, separately. When extending to BERT (SSCL-BERT), we don't add any augmentation methods to construct positives;  Extending to SimCSE (SSCL-SimCSE) means we  utilize dropout masks as the way of building positives.

\paragraph{STS tasks} Table \ref{tab:main_sts} reports the results of methods on 7 STS datasets. From the table, we can observe that: (1) Glove embeddings outperforms BERT$_{base}$, indicating the anisotropy issue has the negative impact of BERT sentence representations; (2) SSCL-BERT$_{base}$ (cls./avg.) surpasses BERT$_{base}$ (cls./avg.) by a large margin (62.42$\%$ vs. 46.70$\%$, 62.56$\%$ vs. 52.54$\%$), showing the effectiveness of our proposed SSCL; (3) SSCL-SimCSE$_{base}$ boosts the model performance of  SimCSE$_{base}$ (77.90$\%$ vs. 76.25$\%$), representing SSCL can easily extend of other contrastive model which can be seen as a plug-and-play framework. Results also prove incorporating negatives  in contrastive learning  is essential for obtaining better sentence representations. Similar results can be observed in the large version of the above models. 
\paragraph{Transfer tasks}
Table \ref{tab:main_transfer} includes the main results on 7 transfer datasets. From the table, we can draw a conclusion that our model SSCL-BERT$_{base}$/SSCL-BERT$_{large}$ outperforms BERT$_{base}$/BERT$_{large}$ on seven datasets, proving the effectiveness of ours. Meanwhile, SSCL-SimCSE$_{base}$/SSCL-SimCSE$_{large}$ also shows a substantial model performance boost when compared with SimCSE$_{base}$/SimCSE$_{large}$. For example, SSCL-SimCSE$_{large}$ improves SimCSE$_{large}$ to 88.88$\%$ (87.17$\%$), suggesting its effectiveness.
\section{Analysis}
\begin{table}[t]
    \begin{center}
    \centering
    \small
    \begin{tabular}{lccc}
    \toprule
        \multirow{2}{*}{Model}  & \textbf{TreeDepth} & \textbf{SentLen} &\textbf{CoordInv} \\
        & (Syntactic) & (Surface) & (Semantic) \\
        \midrule
        BERT & 21  & 67 & 34 \\
    \textbf{SSCL}$^\heartsuit$ & \textbf{23.1}  & \textbf{75.3} & \textbf{42.1} \\
    \midrule
      SimCSE & 24  & 80 & 50 \\
    \textbf{SSCL}$^\clubsuit$ & \textbf{25.3}  & \textbf{88.5} & \textbf{60.18} \\
 
    \bottomrule
    \end{tabular}
    \end{center}
    \caption{
        Probing task performances for each model. \textbf{SSCL}$^\heartsuit$ denotes SSCL  based on BERT, \textbf{SSCL}$^\clubsuit$ represents SSCL based on SimCSE. Concretely, we use the large version of these models.
    }
    \label{tab:probing}
    \vspace{-10pt}
\end{table}

\label{analysis}
In this section, we first conduct qualitative
experiments  via probing tasks to analyse the structural of the resulting representations (Table \ref{tab:probing}), including syntactic, surface and semantic. 
Then, we explore adequate quantitive analysis to verify the effectiveness of SSCL, such as the \textbf{negative sampling strategy}, strengths of SSCL in reducing redundant semantics (\textbf{vector dimension}) and etc. Subsequently, we further provide some discussions on SSCL, like \textbf{chicken-and-egg  issue}.
In the Appendix \ref{more_analysis}, we show the strength of SSCL in fasting convergence speed (Figure \ref{fig:speed}), and conduct discussions: \textbf{whether improvements of resulting model  are indeed from SSCL or just more negatives} (Table \ref{tab:bs_table}).

\subsection{Qualitative Analysis}

\paragraph{Representation Probing}
In this component, we aim to explore the reason behind the effectiveness of the proposed SSCL. Therefore, we conduct some probing tasks to investigate the linguistic structure implicitly learned by our resulting model representations. We directly evaluate each model using three group sentence-level probing tasks: surface task probe for Sentence Length (SentLen), syntactic task probe for the depth of syntactic tree (TreeDepth) and the semantic task probe for coordinated clausal conjuncts (CoordInv). We report the results in Table \ref{tab:probing}, and we can observe our models significantly surpass their original baselines on each task. Specially, SSCL-BERT and SSCL-SimCSE improve the baselines' (BERT and SimCSE) ability of capturing sentence semantic (60.18$\%$ vs. 50$\%$, 42.1$\%$ vs. 34$\%$) and surface (75.3$\%$ vs. 67$\%$, 88.5$\%$ vs. 80$\%$) by a large margin,  which are essential to improve model sentence representations, showing the reason of  ours  perform well on both STS and  Transfer tasks.
\begin{figure*}[t]
    \centering
    \includegraphics[width=0.95\linewidth]{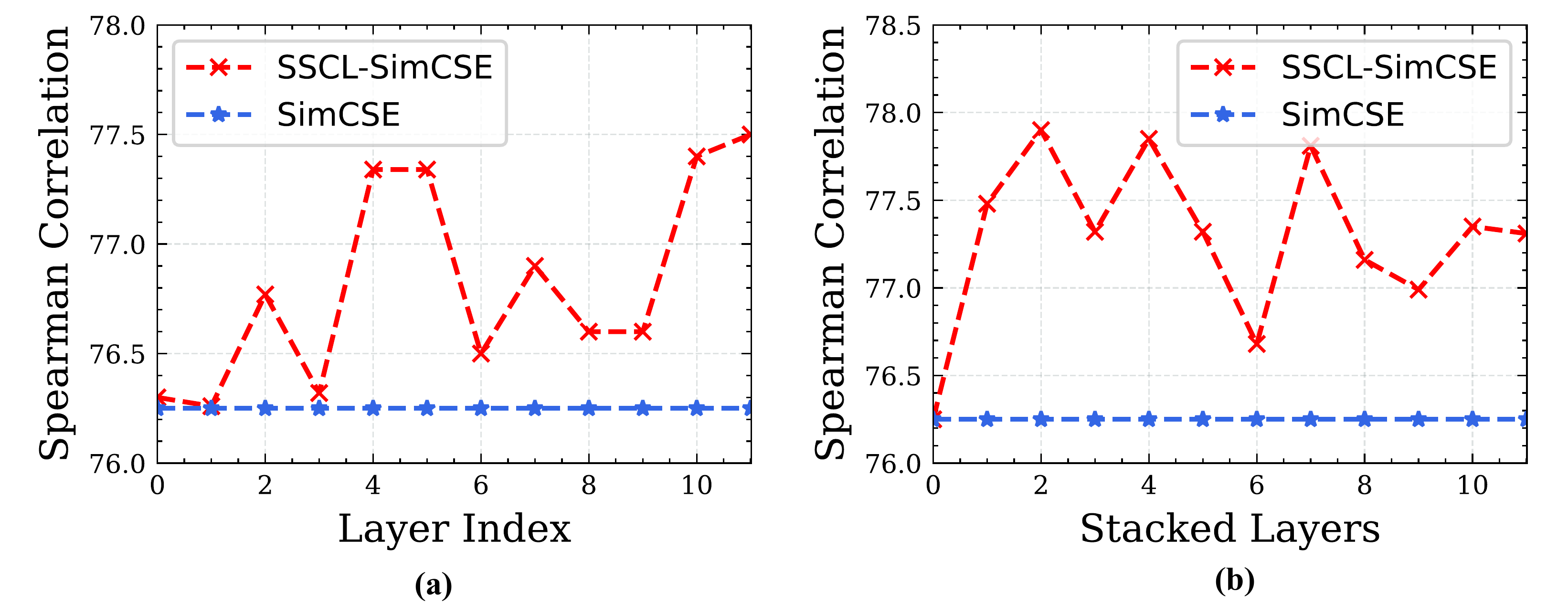}
    \caption{Layer negatives analysis of SSCL. (a) SSCL with different single layers to construct negatives, (b) SSCL with stacked different layers to construct negatives named Progressive SSCL. We report the  average model performances in STS datasets.
    }
    \label{fig:layers}
    \vspace{-5pt}
\end{figure*}

\subsection{Quantitive Analysis}

\paragraph{Negative Sampling Strategy}
From the description in Section \ref{method}, we can raise an intuitive question: Which single layer is most suitable for building negatives in SSCL? Hence, we conduct a series of experiments to verify the effect of intermediate layers with $\{0,1,2,3,4,5,6,7,8,9,10,11\}$, results illustrated in Figure \ref{fig:layers} (a). In the figure, layer-index 0 represents original SimCSE, and layer-index 1-11 represents corresponding transformer layers. We can observe that our model SSCL-SimCSE obtains 
the best result 77.80$\%$ while utilizing 11-th layer representation as  negatives. The reason behind this phenomenon can be explained  that SSCL makes the PLMs more distinguishable between last layer and previous layers, and thus alleviating over-smoothing issues. More specifically, this effect will be more obvious when utilizing 11-th layer representation as negatives, helping the model achieve best result.

\paragraph{Progressive SSCL}
Intuitively, we also can stack several intermediate layers to construct more negatives in our SSCL implement. Thus, we stack previous several layers for building negatives which named Progressive SSCL. We visualize the results in Figure \ref{fig:layers} (b), and the stack transformer layers range 0 to 11. Stacking 0 layer represents original SimCSE, and Stacking 1-11 layers means we stack last 1-11 layers representation to construct negatives. For example, stacking 2 layers represents utilizing 11-th and 10-th transformer layers to form negatives. From the figure, we can draw the following conclusion: (1) Progressive SSCL slightly outperforms SSCL, showing incorporating more negatives can help improve the model performance; (2) 
Progressive SSCL with 2 layers can lead the best model performance (77.90$\%$), indicating using 11-th and 10-th transformer layers to construct negatives can further make the token representations of last layer become more distinguishable.



\paragraph{Vector Dimension}
From the above analysis and experimental results, we can observe SSCL can help the PLMs achieve sufficient sentence-level semantic representations. Therefore, we conduct experiments to verify whether our methods need  high vector dimensions (e.g., 768) to maintain corresponding results. We report  results of BERT, SSCL-BERT and SSCL-SimCSE with different vector dimensions in the Table \ref{tab:dimension}. First, we can observe that BERT performance keeps dropping when word vector dimension reducing, indicating the high vector dimension is essential for maintaining BERT performance.  Then, we also find SSCL-BERT and SSCL-SimCSE still achieve comparable performances with smaller vector dimensions, showing our method can reduce redundant information in the resulting sentence-level representations, and thus lower dimensions is enough for SSCL models obtaining competitive results. It is worthwhile to notice that SSCL-BERT achieves better performances when the vector dimension decreased.  
For example, SSCL-BERT improves the model results from 58.83$\%$ to 62.42$\%$ when vector dimensions reduced from 768  to 256. 

\paragraph{Impact of $\tau$} Intuitively, it is essential to study the sensitivity analysis of the temperature $\tau$ in contrastive learning. Thereafter, we conduct additional experiments to verify the effectiveness of $\tau$ on optimizing the model. We test the model performances with $\tau \in \{0.001, 0.01, 0.05, 0.1\}$. From the Table \ref{tab:temp}, we  observe the different $\tau$ indeed brings performance improvements or drops of both models, and ours achieve best results when $\tau=0.05$.

\subsection{Discussion on SSCL}
\label{discussion_sscl}

\begin{figure}[t]
    \centering
    \includegraphics[width=0.99\linewidth]{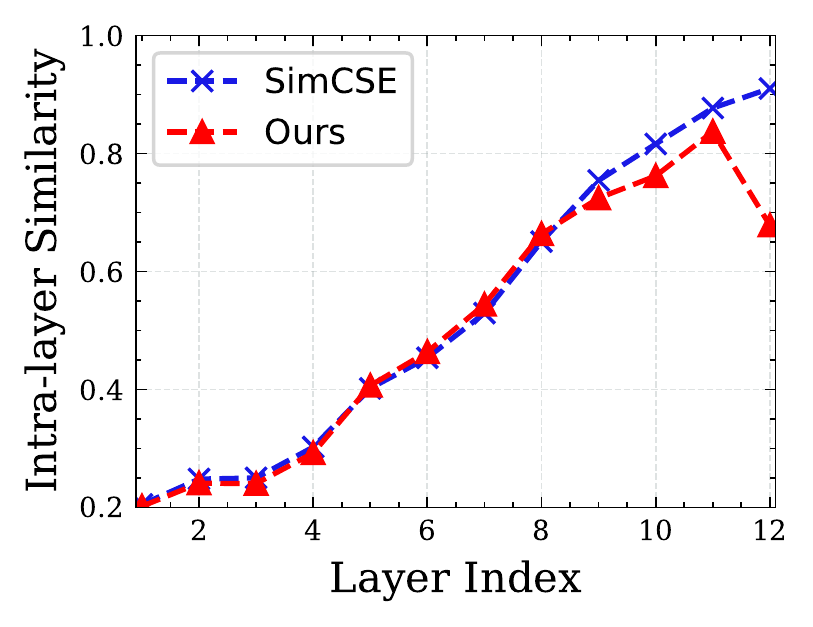}
    \caption{ $\mathrm{TokenSim}$ with different layers computed from SimCSE and Ours (SSCL-SimCSE). We conduct our analysis on STS-B test dataset. In this example, we extend our methods of SimCSE with utilizing the penultimate layer as negatives.
    }
    \label{fig:identical_align}
\end{figure}

\paragraph{Chicken-and-egg issue}
As mentioned in Section \ref{introduction}, our methods effectively alleviate the over-smoothing problem in sentence-level.
In this component, we also utilize $\mathrm{TokSim}$ in Eq.\ref{token_sim} to conduct quantitative analysis to verify whether SSCL could alleviate the over-smoothing problem in \textit{intra-layer} level.
We calculate $\mathrm{TokSim}$  for each  sample from STS-B \cite{cer-etal-2017-semeval} test set with SimCSE and our resulting model  SSCL-SimCSE. For comparison, both models are initialized from BERT stacked with 12 transformer blocks. 
As shown in the Figure \ref{fig:identical_align}, $\mathrm{TokSim}$ is low from the first few layers, showing token representations are highly distinguishable. However, $\mathrm{TokSim}$ becomes higher with layers getting deeper. Concretely,  $\mathrm{TokSim}$ of the last layer from SimCSE is larger than 90$\%$. Thereafter, ours has a obvious $\mathrm{TokSim}$ drop in the last few layers (11 and 12), proving our method alleviates the over-smoothing issue in both sentence level and token level while improving the model performances (Figure \ref{fig:identical_align} (b)). This is because sentence representations are frequently  obtained via adding aggregation methods (e.h., mean pooling and max pooling) over the token representations, resulting  in an entangled relationship \cite{DBLP:conf/emnlp/MohebbiMP21}. Therefore, alleviating over-smoothing in sentence representation could eliminate over-smoothing at token-level to some extent.

\begin{table}[t]
    \begin{center}
    \centering
    \small
    \begin{tabular}{lcc}
    \toprule
        \textbf{Model} & \textbf{Dimension} & \textbf{Avg. STS} \\
    \midrule
      \multirow{3}{*}{BERT} & 128  & 39.24  \\
                  & 256 & 43.22 \\
                  & 768 &  \textbf{46.7} \\
                  \midrule
\multirow{3}{*}{SSCL-BERT} & 128 & 61.3 \\
                  & 256 & \textbf{62.42} \\
                  & 768 & 58.83 \\
                  \midrule
 \multirow{3}{*}{SSCL-SimCSE} & 128 & 76.53  \\
                  & 256 & 77.97 \\
                  & 768 & \textbf{77.90}\\
    \bottomrule
    \end{tabular}
    \end{center}
    \caption{
        Ablation studies of the vector dimension based on the development sets using BERT. We utilize the same simply linear projection head to transfer the vector dimensions. Here, the Spearman's Correlation is employed  as the evaluation metric.
    }
    \label{tab:dimension}
\end{table}
\begin{table}[]
    \centering
    \begin{tabular}{lcccc}
    \toprule
    \multirow{2}{*}{Model} & \multicolumn{4}{c}{$\tau$} \\
    \cmidrule{2-5}
         & 0.001 & 0.01 & 0.05 & 0.1 \\
      \midrule
       SimCSE  & 74.82 & 75.33 &\textbf{76.25}&72.24\\
       SSCL$^\clubsuit$&75.77 &77.40 & \textbf{77.90} &74.12 \\
       \bottomrule
    \end{tabular}
    \caption{Model performances with different $\tau$ during training. We report average results of SimCSE and SSCL-SimCSE on STS tasks. SSCL$^\clubsuit$ denotes SSCL-SimCSE.}
    \label{tab:temp}
    \vspace{-5pt}
\end{table}

\label{case_study}
\begin{figure}[!t]
\centering
\includegraphics[width=1\linewidth]{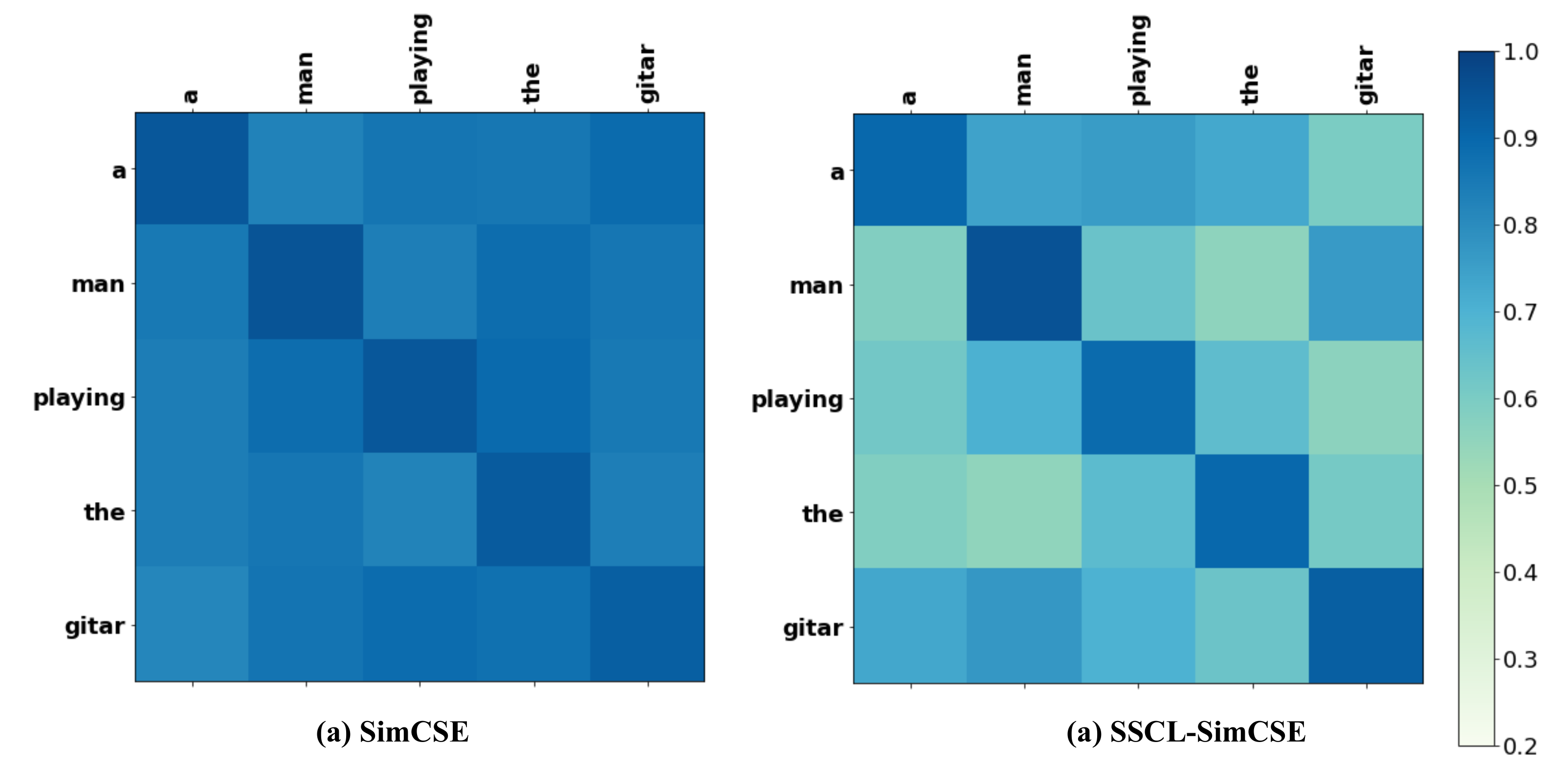}
\caption{Token representation cosine similarity  matrices from SimCSE and SSCL-SimCSE. Representations taken from the last layers of these models. }
\label{fig:attention}
\end{figure}

\paragraph{Visualization} As shown in the Figure \ref{fig:attention} (a), we showcase the token representation similarities produced by SimCSE \cite{DBLP:journals/corr/abs-2104-08821}. Obviously, we can observe each token representation in the sentence is very close to each other. Nevertheless, the token representations within the same sentence should be discriminative even if the sentence structure is simple in the ideal setting (as shown in the Figure \ref{fig:attention} (b)). 
As aforementioned, such high similar token representations may confuse the model to capture global and reasonable sentence-level understanding, leading to sub-optimized sentence representations. Nevertheless, our SSCL-SimCSE   can alleviate this problem from the \textit{inter-layer} perspective while making  token representations in the sentence more discriminative, as seen in Figure \ref{fig:attention} (b).

\section{Conclusion}
\label{conclusion}
In this paper, we explore the over-smoothing problem in unsupervised sentence representation. Then, we propose a simple yet effective method named SSCL, which constructs negatives from PLMs intermediate layers to alleviate this problem, leading better sentence representations. The proposed SSCL can easily be extended to other state-of-the-art methods, which can be seen as a plug-and-play contrastive framework.  Experiments on seven STS datasets and seven Transfer datasets prove the effectiveness of our proposed method. And  qualitative analysis indicates our method improves the resulting model's ability of capturing the semantic and surface. Also quantitative analysis shows the proposed SSCL not only reduces redundant semantics but also fasts the convergence speed. As an extension of our future work, we will explore other methods to improve the unsupervised sentence representation quality.
\newpage
\section*{Broader Impacts}
The main contributions of this paper are towards tackling over-smoothing issue for learning unsupervised sentence representation. The proposed approach is fairly basic and may simply be extended to improve the performance of other state-of-the-art models. More broadly, we anticipate that the central idea of this study will provide insights to other research communities seeking to improve sentence representation in an unsupervised setting. Admittedly, the proposed strategies are restricted with unsupervised training, and biases in the training corpus also may influence the performance of the resulting model.
These concerns warrant further research and consideration when utilizing this work to build unsupervised  retrieval systems.


\bibliography{anthology,custom}
\bibliographystyle{acl_natbib}
\clearpage
\appendix


\section{Related Work}
\label{related_work}
\subsection{Unsupervised Sentence Representation }
Unsupervised sentence representation learning has gained lots of attention, which is  considered to be one of the most promising areas in natural language understanding. Thanks to remarkable results achieved by PLMs, quite a few works \cite{DBLP:journals/corr/abs-1810-04805,DBLP:conf/iclr/LanCGGSS20} tended  to  directly use the output of PLMs, obtaining the sentence-level representation via \texttt{[CLS]} token-based representation or leveraging pooling methods (e.g., mean-pooling and max-pooling). Recently,  some works \cite{li-etal-2020-sentence,su2021whitening,DBLP:journals/corr/abs-2202-08625} found that there are anistropy and over-smoothing problems in BERT \cite{DBLP:journals/corr/abs-1810-04805} representations. Facing these challenges, \citet{su2021whitening} introduced whitening methods to obtain isotropic sentence embedding distribution. More recently, \citet{DBLP:journals/corr/abs-2202-08625} proposed to alleviate over-smoothing problem via graph fusion methods. In this paper, we design a novel and simple approach to improve the quality of sentence representations, making them more uniform while alleviating the  over-smoothing problem from a new perspective.

\subsection{Contrastive Learning} During the past few years, contrastive learning \cite{DBLP:conf/cvpr/HadsellCL06} has been proved as an extremely promising approach to build on learning effective representations in different contexts of deep learning \cite{chen2021good,chen-etal-2022-bridging,DBLP:journals/corr/abs-2104-08821,DBLP:journals/corr/abs-2106-02182,you2021self,you2021mrd,chen2023bridge}. Concretely, contrastive learning objective aims at
 pulling together semantically close positive samples (short for positives) in a  semantic space, and pushing apart negative samples (short for negatives). In the context of learning unsupervised sentence representation, \citet{DBLP:journals/corr/abs-2012-15466} proposed leveraging several sentence-level augmentation strategies to construct positives, obtaining a noise-invariant representation. Recently,  \citet{DBLP:journals/corr/abs-2104-08821} designed a simple method named SimCSE for constructing positives into contrastive learning via using  dropout \cite{DBLP:journals/jmlr/SrivastavaHKSS14} as noise. In detail, \citet{DBLP:journals/corr/abs-2104-08821} passed the each sentence into the PLMs twice and obtained positives by applying random dropout masks in the representations from last layer of PLMs.
 Subsequently, \citet{DBLP:journals/corr/abs-2201-05979} extended of SimCSE to formulate a new contrastive method called MixCSE, which continually constructing hard negatives via mixing both positives and negatives. However, it is  still limited to the specific framework. In this paper, we focus on mining hard negatives for learning unsupervised sentence representation without complex data augmentation methods and not limited to some specific frameworks. Accordingly, we propose SSCL, a plug-and-play framework, which can be extended to various state-of-the-art models. 
\label{speed}
\begin{figure}[t]
    \centering
    \includegraphics[width=1\linewidth]{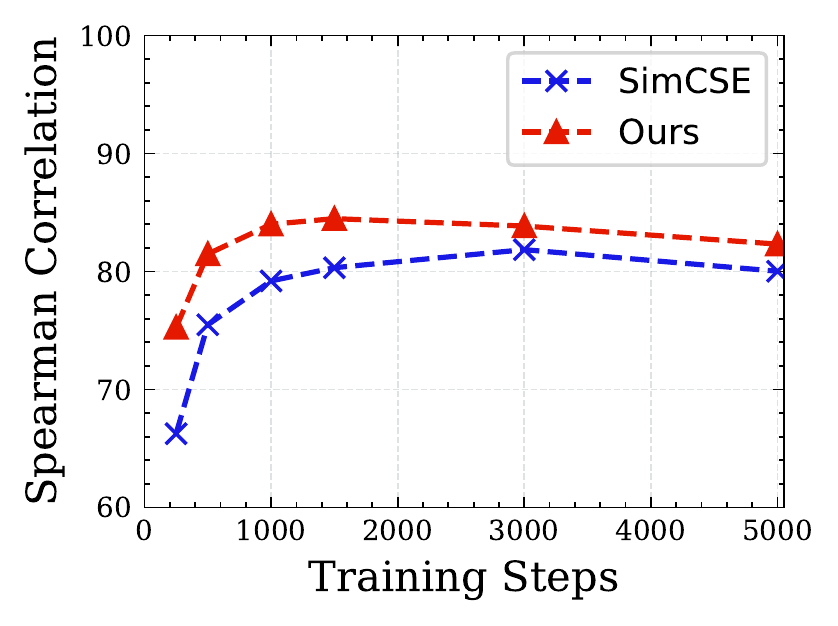}
    \caption{Convergence speed of SimCSE and SSCL-SimCSE. We evaluate the above models on STS-B development set.
    }
    \label{fig:speed}
    \vspace{-5pt}
\end{figure}
\section{More Analysis}
\label{more_analysis}
\subsection{Convergence Speed} 
Moreover, we report the convergence speed of SimCSE and our resulting model: SSCL-SimCSE in  the Figure \ref{fig:speed}. From the figure, we can observe that SimCSE and SSCL-SimCSE both obtain their best results before the training ends. And SSCL-SimCSE  manages to maintain an absolute lead of 5$\%$-15$\%$ over SimCSE  during the early stage of training, showing our methods not only speed the training time and achieves superior performances.  Concretely, SSCL-SimCSE achieves its best performances with only 1500 steps iteration. That is, our model can fast the convergence speed greatly, and thus, save the time cost.

\subsection{Discussion on More Negatives}
\begin{table*}[]

    \centering
    \small
    \begin{tabular}{lccccccccc}
    \toprule
       $\mathbf{Model}$ & $\mathbf{BS}$ &  $\mathbf{STS12}$ & $\mathbf{STS13}$ & $\mathbf{STS14}$ & $\mathbf{STS15}$ & $\mathbf{STS16}$ & $\textbf{STS-B}$ & $\textbf{SICK-R}$ & $\mathbf{Avg.}$ \\
       \midrule
\multirow{2}{*}{BERT~(cls.)} & 64 & 29.70&	49.38&	39.67&	56.03&	56.19&	43.87&	52.06&	46.70\\
& 128 & 31.05&	49.96&	40.54&	57.68&	57.05&	45.99&	52.95&	47.89\\
\midrule
$\textbf{SSCL-BERT}$~(cls.) & 64 & 49.21&	\textbf{67.59}&	\textbf{58.96}&	\textbf{69.94}&	68.00&	62.87&	60.43&	62.42\\

       \midrule
  \multirow{2}{*}{SimCSE}   & 64 & 68.40&	82.41 &	74.38&	80.91&	78.56&	76.85&	$\mathbf{72.23}$&	76.25\\
     & 128 & 69.49&	82.75 &	74.98&	81.09&	77.89&	77.15&	70.06&	76.21\\
   \midrule
         $\textbf{SSCL-SimCSE}$ & 64&$\mathbf{71.68}$&	$\mathbf{83.50}$ &	$\mathbf{76.42}$&	$\mathbf{83.46}$&78.39&	$\mathbf{79.03}$&71.76&	$\mathbf{77.90}$ \\
         \bottomrule
    \end{tabular}
    \caption{Model performances under different batch size.}
    \label{tab:bs_table}
\end{table*}
As illustrated in Eq.\ref{eq:simhne}, our SSCL enlarges the size of  mini-batch negatives from N pairs to 2N pairs. Intuitively, there is a question: whether the improvements of the resulting model are from SSCL? Or the model can achieve such results via just enlarging the batch size to get more in-batch negatives. To answer this question, we conduct additional experiments, as shown in Table \ref{tab:bs_table}. When enlarging the batch size from 64 to 128, SimCSE still achieves comparable performances rather than obtaining obvious improvements like SSCL-SimCSE. In other words, simply expanding in-bath negatives can not effectively lead to better sentence representations, that is, the performance boost of SSCL-simCSE indeed comes from our method.



\end{document}